\documentclass{article}

\usepackage{microtype}
\usepackage{graphicx}
\usepackage{subfigure}
\usepackage{booktabs} % for professional tables
\usepackage{amsmath, amssymb}
\usepackage{natbib}
\usepackage{sidecap}
\usepackage{listings}
\usepackage{color}
\usepackage{makecell}
\usepackage{skak}

\usepackage{hyperref}

\newcommand{\RNN}{{\sc rnn}}      % baseline recurrent net
\newcommand{\RNNSR}{{\sc rnn+sr}} % recurrent net with state reification
\newcommand{\RNNA}{{\sc rnn+}}    % recurrent net with additional layers for attractor net, but no state reif.
\newcommand{\CNN}{{\sc cnn}}      % baseline conv net
\newcommand{\CNNSR}{{\sc cnn+sr}} % conv net with state reification
\newcommand{\CNNA}{{\sc cnn+}}    % conv net with additional layers for DAE, but no state reification
\renewcommand\epsilon\varepsilon

\newcommand\hood{\mathcal{S}}

\usepackage{tikz,pgfplots}
\usetikzlibrary{backgrounds}
\usetikzlibrary{calc}
\usetikzlibrary{fit}
\usetikzlibrary{positioning}

\usepackage[accepted]{icml2019}

\icmltitlerunning{State Reification}

\begin{document}

%independent ? disentangled ?

\twocolumn[
\icmltitle{State-Reification Networks: Improving Generalization\\by Modeling the Distribution of Hidden Representations}



\icmlsetsymbol{equal}{*}

\begin{icmlauthorlist}
\icmlauthor{Alex Lamb}{udem,equal}
\icmlauthor{Jonathan Binas}{udem,mila}
\icmlauthor{Anirudh Goyal}{udem}
\icmlauthor{Sandeep Subramanian}{mila}
\icmlauthor{Ioannis Mitliagkas}{udem}
\icmlauthor{Denis Kazakov}{colo}
\icmlauthor{Yoshua Bengio}{mila}
\icmlauthor{Michael C. Mozer}{colo,goog,equal}
\end{icmlauthorlist}

\icmlaffiliation{goog}{Presently at Google Research, Mountain View, CA}
\icmlaffiliation{udem}{Universit\'e de Montr\'eal, Montr\'eal, Quebec}
\icmlaffiliation{mila}{Montr\'eal Institute for Learning Algorithms, Montr\'eal, Quebec}
\icmlaffiliation{colo}{University of Colorado, Boulder, CO}

\icmlcorrespondingauthor{Alex Lamb}{lambalex@iro.umontreal.ca}
\icmlcorrespondingauthor{Michael C. Mozer}{mcmozer@google.com}

\icmlkeywords{Machine Learning, ICML, neural networks, deep learning, state reification, generalization, adversarial robustness, denoising autoencoder, attractor network}

\vskip 0.3in
]



\printAffiliationsAndNotice{\icmlEqualContribution} % otherwise use the standard text.

\begin{abstract}

% MM's stab at an abstract:
Machine learning promises methods that generalize well from finite labeled data. However, the brittleness of existing neural net approaches is revealed by notable failures, such as the existence of adversarial examples that are misclassified despite being nearly identical to a training example, or the inability of recurrent sequence-processing nets to stay on track without teacher forcing. We introduce a method, which we refer to as \emph{state reification}, that involves modeling the distribution of hidden states over the training data and then projecting hidden states observed during testing toward this distribution. Our intuition is that if the network can remain in a familiar manifold of hidden space, subsequent layers of the net should be well trained to respond appropriately. We show that this state-reification method helps neural nets to generalize better, especially when labeled data are sparse, and also helps overcome the challenge of achieving robust generalization with adversarial training.

\end{abstract}

\section{Introduction}

\begin{figure}[tb]
\centering
\includegraphics[width=3.25in]{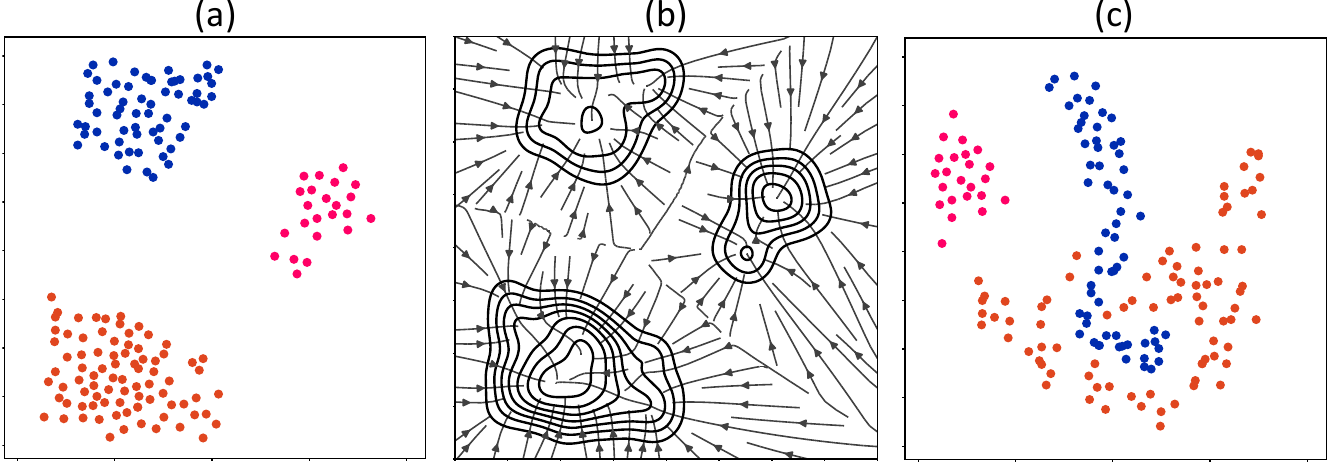}
\caption{(a) A distribution of hidden states, with class label indicated by color. (b) State reification dynamics to map hidden states toward regions of higher density. (c) A distribution of input states, showing poorly separated classes, making it not suitable for reification.}
\label{fig:basic}
\end{figure}

The fundamental objective of machine learning is to build models of complex data. By abstracting from the data, models are typically more useful for domain understanding and prediction than are the raw data. This substitution of a model in place of the data is a form of \emph{reification}. In this article, we argue that reification of data has similar value even when the data originate from within the model, i.e., its latent states. We propose a recursive model-within-a-model that reifies internal states in a neural network, leading to robustness and improved generalization.
%%JB: shorter, one could say that we're proposing a new type of regularizer (but i'm not sure this is stated in the text)

Our proposed method, which we call \emph{state reification}, is based on the idea that it is possible to model the distribution of the hidden states over the training data, and then map less likely states toward more likely states. Because the network has experienced these latter states frequently during training, we would expect to obtain better generalization from them.
To offer an intuition, consider a simple task: training a recurrent net to output the parity of a stream of binary digits. The ideal internal state for solving this task is discrete, yet deep neural networks have continuous activations. Consequently, when evaluated on long test sequences, the continuous dynamics may cause the net to wander from the ideal states. State reification will map these rogue states back to the values observed during training, leading to dramatically better generalization.

Our approach stems from the observation that latent states, such as hidden representations, tend to lie on one or more manifolds. Figure \ref{fig:basic}a depicts a hidden representation of a training set in a classification task, with class label indicated by color.  States within the manifold are `familiar' in the sense that subsequent layers of the net have been tuned to process them. However, states lying outside the manifold are potentially problematic; they are not reached given the distribution of training inputs, and therefore the model's extrapolatory response may be unreliable.

In this work, we explore an approach in which we construct a model-within-a-model that implicitly encodes the distribution of states in the latent space and then projects states from off-manifold regions back to the manifold, where the network is likely to perform robustly (Figure \ref{fig:basic}b). We argue that this projection operation serves as a useful inductive bias during training that restricts the state space and  induces a clustering of states. Not only does it boost generalization performance,
but it also makes networks less sensitive to adversarial input perturbations, which tend to throw the state off the training manifold. Explicit detection of off-manifold states has proven useful for adversarial robustness and detecting out-of-distribution samples \cite{Carraraetal2018,LeeLeeLeeShin2017,LeeLeeLeeShin2018}, as has incorporating losses to shape the manifolds \cite{PangDuDongZhu2018}. Without explicitly addressing manifolds, \citet{Liao2016} proposed a clustering-based regularization objective to encourage parsimonious representations.
We present a general method that goes beyond clustering states and identifying off-manifold states by projecting states back to the manifold. Like \citet{Liao2016}, our method can be applied to any architecture or any layer of a network.

One could in-principle reify off-manifold inputs rather than off-manifold hidden states. A large body of literature exists on this topic, from early work achieving noise robustness via preprocessing stages that estimate and filter noise from an input signal \cite{Boll1979} to more recent work in machine learning involving loss functions to achieve invariance to task-irrelevant perturbations in the input \cite{Simard1992,Zheng2016}. However, there are two reasons to prefer reification of hidden states. First, distinct semantic classes are typically more intertwined in the input space than in the hidden space (Figure \ref{fig:basic}c), and the manifolds are therefore simpler and smoother in an abstract space with simpler statistical structure.
Second, state reification should have particular value in recurrent nets in which steps off manifold may compound as the hidden state evolves over a sequence.
% WOULD LIKE TO ADD THIS IF THERE IS SPACE. PROBABLY TO THE INTRO. TO ADD THIS, WE'D NEED TO ADD THE ATTRACTOR LANDSCAPE FOR THE INPUT SPACE. PROBABLY BEST TO DE-EMPHASIZE THE HIDDEN VS. INPUT SPACE, BECAUSE THAT MAKES MORE SENSE FOR ADVERSARIAL GENERALIZATION THAN RECURRENT NET GENERALIZATION.
%As illustrated in Figure~\ref{fig:illustration_fortnet}, we hypothesize that more abstract representations associated with deeper networks are easier to denoise because the transformed data manifolds are flatter. The flattening of data manifolds in the deeper layers of a neural network was first noted experimentally by~\citet{Bengio-et-al-ICML2013-small}. We provide experimental support for these claims in Section~\ref{sec:experiments}.

We also briefly note that State Reification closely builds on the ideas presented in State-Denoising RNNs \citep{mozer2018sdrnn} and Fortified Networks \citep{lamb2018fortified}.

\section{State Reification}
To show the robustness of our underlying insight, we describe two distinct but related mechanisms for state reification: denoising autoencoders and attractor networks.
%%JB 1.22: It might be good to point out that this is not explicitly about DAEs or attractors, but pretty much any kind of (compressing) representation learning method could be used.
%%MM 1/23: good idea. let's do this in the discussion

\subsection{Denoising Autoencoders}

\emph{Denoising autoencoders} (DAEs) are neural networks that map a noise-corrupted version of vector $x$ to a clean version of $x$.  This approach has been widely used for feature learning and generative modeling in deep learning~\citep{Bengio-Courville-Vincent-TPAMI2013}.  More formally, denoising autoencoders are trained to minimize a reconstruction error or negative log-likelihood of generating the clean $x$. For example, with Gaussian log-likelihood of the clean vector given the corrupted vector, the reconstruction loss for data set $\mathbf{x}=\{ x^{(1)},~\ldots, x^{(N)}\}$ is
\begin{equation}
\mathcal{L}_\mathrm{rec}(\mathbf{x}) = \frac{1}{N} \sum_{n=1}^N
   \left(
      \left\Vert r_\theta \left(x^{(n)} + a^{(n)} \right) - x^{(n)} \right\Vert _2^2
   \right),
  \label{eqn:rec_loss}
\end{equation}
where $r_\theta$ is the learned denoising function and $a^{(n)}~\sim~\mathbb{N}(\boldsymbol{0},\sigma^2\boldsymbol{\mathrm{I}})$ is a Gaussian noise vector.

Given loss $\mathcal{L}_\mathrm{rec}$ and Gaussian corruption, a well-trained denoising autoencoder's reconstruction vector is proportional to the gradient of the log-density \citep{alain2012dae}:
\begin{equation}
    \frac{r_{\sigma}(x) - x}{{\sigma^2}} \rightarrow \frac{\partial \log p(x)}{\partial x}
    \hspace{1em} \textrm{as} \hspace{1em} {\sigma} \rightarrow 0. \label{eqn:rx-x-trick}
\end{equation}

The theory of \citet{alain2012dae} establishes that the reconstruction vectors from a well-trained denoising autoencoder form a vector field which points in the direction of the data manifold. However, this result is not guaranteed for points distant from the manifold, as these points are rarely sampled during training. In practice, denoising autoencoders are trained with not just tiny noise levels but also with large noise levels, which blurs the data distribution as seen by the learner but makes the network learn a useful vector field even far from the data. 

\subsection{Attractor Networks}

DAEs can be applied iteratively by cycling the output back to the input. A related but more principled approach is an \emph{attractor network} (AN), which is essentially a DAE with recurrent connections within the hidden layer that results in a discrete-time nonlinear dynamical system with attractor manifolds, achieving trajectories like those shown in Figure~\ref{fig:basic}b.
Attractor nets have a long history starting with the seminal work of
\cite{Hopfield1982} that was partly responsible for the 1980s wave of
excitement in neural networks.  
We adopt Koiran's (\citeyear{Koiran1994}) framework, which dovetails with the
standard deep learning assumption of synchronous updates on continuous-valued
neurons. Koiran shows that a hidden layer with symmetric weights, nonnegative self-connections, and a bounded nonlinearity that is continuous and strictly increasing except at the extrema (e.g., tanh), the network asymptotically
converges over iterations to a fixed point or limit cycle of length 2.
Although the AN is recurrent, its training barely suffers from vanishing gradients \cite{Bengio1994,hochreiter1998} because the input projects to the
hidden layer at each iteration, acting as a type of skip connection. Technical
details are presented in the supplementary materials.

\subsection{Incorporating State Reification Into a Model}

\definecolor{nnin}{RGB}{112,48,160}
\definecolor{nnout}{RGB}{192,0,0}
\definecolor{nnattin}{RGB}{0,112,192}
\definecolor{nnattout}{RGB}{180,153,0} % {229,172,0} % changed for legibility {255,192,0}
\definecolor{nnattrnet}{RGB}{0,176,80}

\begin{figure}[bt]
\centering
\includegraphics[width=3.25in]{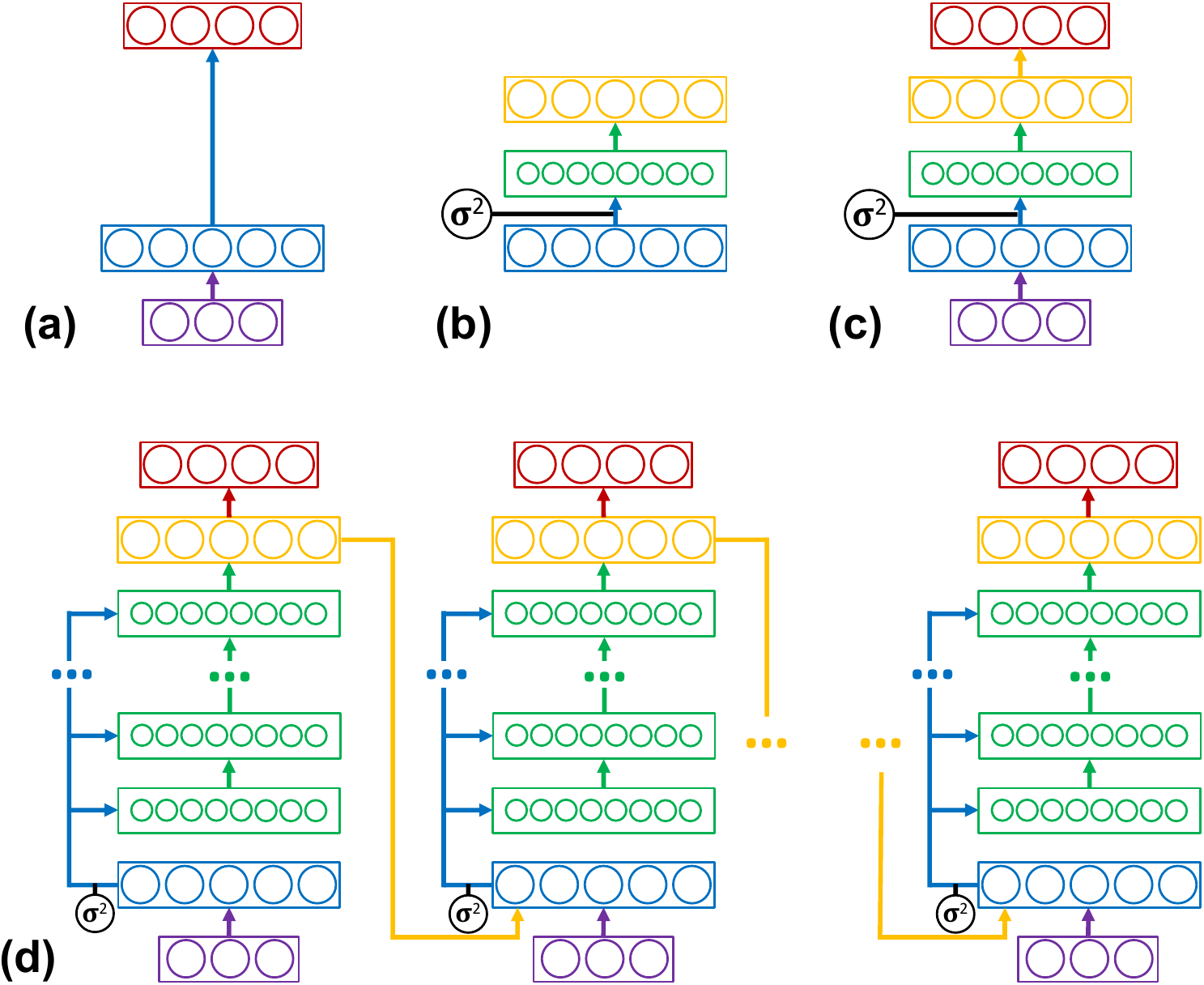}
\caption{(a) A network that performs some \textcolor{nnin}{input}-\textcolor{nnout}{output} mapping task with one intervening \textcolor{nnattin}{hidden} layer. (b) A \textcolor{nnattrnet}{DAE} that produces a \textcolor{nnattout}{reified output}. (c) Integrating the two architectures to perform state reification on the \textcolor{nnattin}{hidden} state. (d) 
A recurrent sequence processing architecture, unrolled in time horizontally, with an \textcolor{nnattrnet}{attractor net}---unrolled vertically---reifying the \textcolor{nnattin}{hidden} state.}
\label{fig:arch}
\end{figure}

We incorporate state reification within a neural net's internal layers to transform the representation toward the training-data manifold.
For example, Figure~\ref{fig:arch}a shows a feedforward net that maps \textcolor{nnin}{inputs} to \textcolor{nnout}{outputs} through a \textcolor{nnattin}{hidden} layer. Figure~\ref{fig:arch}b shows a DAE with one \textcolor{nnattrnet}{internal} layer, producing a \textcolor{nnattout}{reified output}. Figure~\ref{fig:arch}c integrates the feedforward net and DAE to reify the hidden state of the feedforward net. Figure~\ref{fig:arch}d shows a more elaborate architecture, with a recurrent sequence-processing net integrated with an attractor net. Each column denotes a single time step with a corresponding \textcolor{nnin}{input} and \textcolor{nnout}{output}. The \textcolor{nnattin}{hidden state} is denoised by an \textcolor{nnattrnet}{attractor net}, unrolled vertically, yielding a \textcolor{nnattout}{reified state} which is combined with the next \textcolor{nnin}{input} to determine the next \textcolor{nnattin}{hidden state}.
Intuitively, the method aims to regularize the hidden representations by projecting activations to the training-data manifold through the application of a DAE or attractor net.

To train the simple integrated model (Figure~\ref{fig:arch}c), training data are processed in mini-batches,
%$\{ (x^{(1)},y^{(1)}),~\ldots,~(x^{(N)},y^{(N)})\}$, and the loss 
and the loss per $(x,y)$ example is:
\begin{equation}
%\mathcal{L} = \frac{1}{N} \sum_{n=1}^N \mathcal{L}_\mathrm{task}(x^{(n)},y^{(n)}) + \lambda_{\mathrm{rec}}~\mathcal{L}_\mathrm{rec} (h^{(n)})
\mathcal{L} = \mathcal{L}_\mathrm{task}(x,y) + \lambda_{\mathrm{rec}}~\mathcal{L}_\mathrm{rec} (h)
\label{eqn:comb_loss}
\end{equation}
is minimized, where $\mathcal{L}_\mathrm{rec}$ (Equation~\ref{eqn:rec_loss}) is applied to the hidden state, $h$, $\mathcal{L}_\mathrm{task}$ is a primary-task loss, and the coefficient $\lambda_\mathrm{rec} >0$ controls the contribution of reification. This approach allows us in principle to reify multiple hidden layers at once, each with its own $\mathcal{L}_\mathrm{rec}$ loss. In the next sections, we present results for two related applications: obtaining robust generalization to out-of-sample cases in sequential tasks, and obtaining robustness to standard adversarial attacks in feedforward nets. We use slightly different training procedures for each application, due to the different goals. For improving test-set generalization, we train only the reifier (DAE or
AN) weights on $\mathcal{L}_\mathrm{rec}$, all weights on $\mathcal{L}_\mathrm{task}$, and we set the noise level, $\sigma^2=0$, for evaluation. For adversarial robustness, we train all weights on the joint loss, and perform simulations with and without the noise during evaluation; we also incorporate additional adversarial loss terms that are duals to $\mathcal{L}_\mathrm{rec}$ and $\mathcal{L}_\mathrm{task}$, to be described shortly. 

\section{Experiments}

We demonstrate the effectiveness of state reification on three classes of problems: sequence classification in a data-limited training environment, generation of long sequences, and adversarial perturbations in image processing.  

\subsection{Recurrent Networks for Sequence Classification}
\label{sec:symbolic}
Our first experiments involve symbolic sequence-classification tasks using recurrent networks like that in
Figure~\ref{fig:arch}d, where state reification is performed with attractor dynamics. We chose symbolic tasks---tasks with discrete inputs, and input-output mappings that can be characterized in terms of rules---because symbolic tasks have always been a challenge for continuous neural networks \citep{CravenShavlik1993}.

%% TURN these subsubsections into paragraph headings if we're tight on space.
\subsubsection{Parity}
%\underline{\bf Parity}. 
We studied a streamed \emph{parity} task in which 10 binary inputs are presented in sequence and the target output following the last sequence element is $1$ if an odd number of $1$s is present in the input or $0$ otherwise. The architecture has $10$ hidden units, $20$ attractor units, and a single input and a single output. We experimented with both tanh and GRU hidden units. We trained the attractor net with $\sigma=.5$ and ran it for exactly 15 iterations (more than sufficient to converge). Models were trained on 256 randomly selected binary sequences. Two distinct test sets were used to evaluate models: one consisted of the held-out 768 binary sequences, and a second test set consisted of three copies of each of the 256 training sequences with additive uniform $[-0.1,+0.1]$ input noise. We performed one hundred replications of a baseline architecture (\emph{\RNN}), an architecture with the additional layers to implement attractor dynamics but trained solely on  $\mathcal{L}_\textrm{task}$ (\emph{\RNNA}, the '+' indicating the additional hardware), and an architecture with state reification (\emph{\RNNSR}). Other details of this and subsequent simulations are presented in the Supplementary Materials.

\begin{figure}[bt]
\begin{center}
   \includegraphics[width=3.25in]{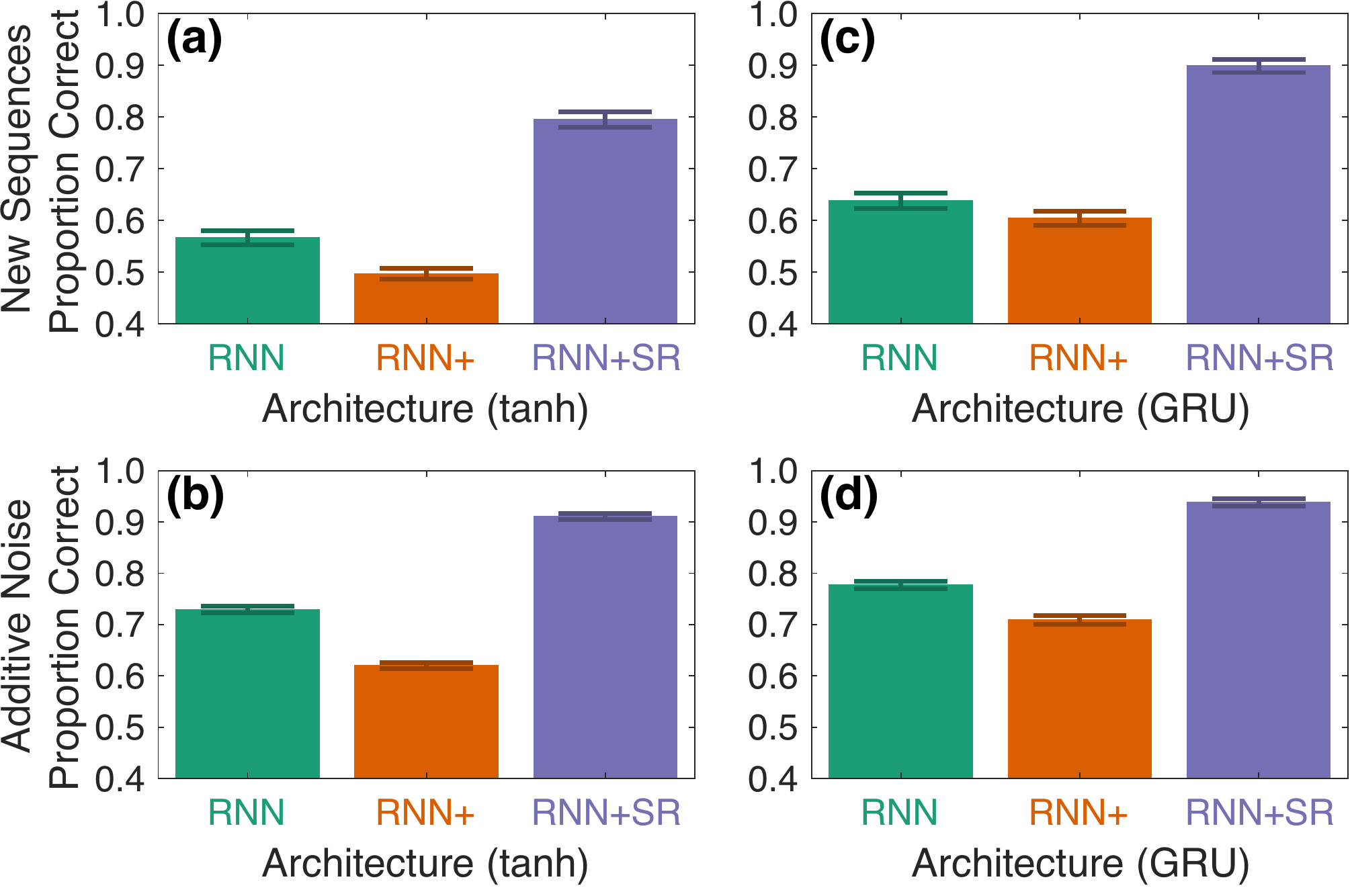} 
   \end{center}
   \caption{Parity simulations. Top row shows generalization performance on novel binary sequences; bottom row shows performance on trained sequences with additive noise. Unless otherwise noted, the simulations of with state reification use $\sigma=0.5$, 15 attractor iterations, and $L_2$ regularization (a.k.a. weight decay) 0.0. Error bars indicate $\pm1$ SEM, based on a correction for confidence intervals with matched comparisons \citep{MassonLoftus2003}.}
   \label{fig:parity}
\end{figure}

Figure~\ref{fig:parity}a shows relative performance on the held-out sequences by the \RNN, \RNNA, and \RNNSR\ with a tanh hidden layer. Figure~\ref{fig:parity}b shows the same pattern of results for the noisy test sequences. \RNNSR\ significantly outperforms both the \RNN\ and the \RNNA: it generalizes better to novel sequences and is better at ignoring additive noise in test cases, although such noise was absent from training. Figures~\ref{fig:parity}c,d show similar results for models with a GRU hidden layer. Absolute performance improves for all three recurrent net variants with GRUs versus tanh hidden units, but the relative pattern of performance is unchanged. Note that the improvement due to denoising the hidden state (i.e., \RNNSR\ versus \RNN\ for both tanh and GRU architectures) is much larger than the improvement due to switching hidden unit type (i.e., \RNN\ with GRU vs. tanh hidden), and that the use of GRUs---and the equivalent LSTM---is viewed as a critical innovation in deep learning.

In principle, parity should be performed more robustly if a system has a highly restricted state space. Ideally, the state space would itself be binary, indicating whether the number of inputs thus far is even or odd. Such a restricted representation should force better generalization. Indeed, quantizing the hidden activation space for all sequence steps of the test set, we obtain a lower entropy for the tanh \RNNSR\ (3.70, standard error .06) than for the tanh \RNN\ (4.03, standard error .05). However, what is surprising  about this simulation is that gradient-based procedures could learn such a restricted representation, especially when two orthogonal losses compete with each other during training. The competing goals are clearly beneficial, as \RNNA\ and \RNNSR\ share the same architecture and differ only in the addition of the denoising loss.

The noise being suppressed during training is neither input noise nor label noise; it is noise in the internal state due to weights that have not yet been adapted to the task. Nonetheless, denoising internal state during training appears to help the model overcome input noise and generalize better.

\subsubsection{Majority Task}

We next studied a \emph{majority} task in which the input is a binary sequence and the target output is $1$ if a majority of  inputs are 1, or $0$ otherwise. We trained networks on 100 distinct randomly drawn fixed-length sequences, for length $l \in \{11,17,23,29,35\}$. We performed 100 replications for each $l$ and each model. We ensured that runs of the various models were matched using the same weight initialization and the same training and test sets. All models had $10$ tanh hidden units, $20$ attractor units, $\sigma=.25$.

%\begin{SCfigure}[1.11111][bt]%[100]%[bt]
\begin{figure}[bt]
    \begin{center}
    \includegraphics[width=3.25in]{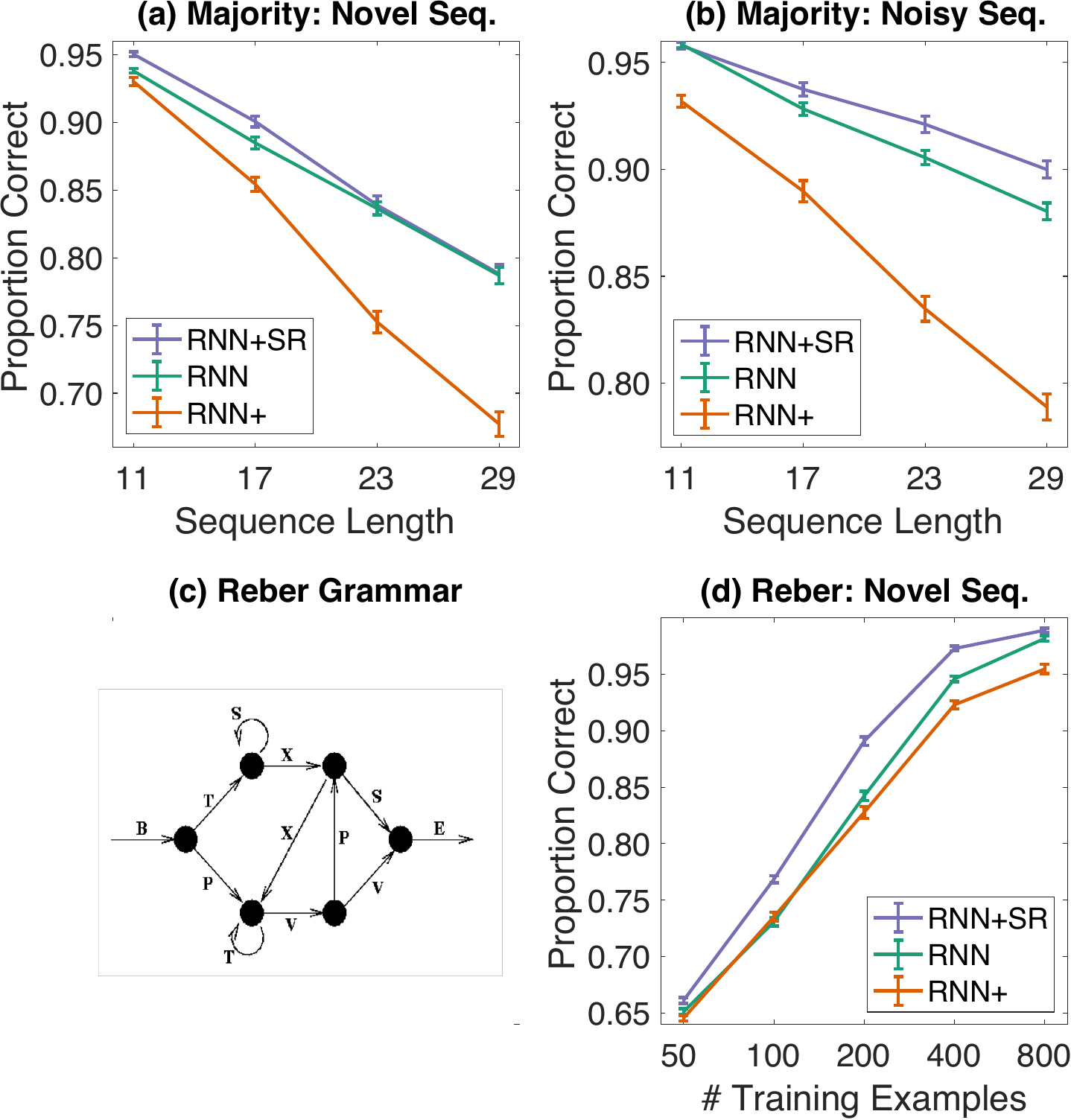} 
    \end{center}
    \caption{Simulation results on majority task with (a) novel and (b) noisy sequences. (c) Reber grammar. (d) Simulation results on Reber grammar. Error bars indicate $\pm1$ SEM, based on a correction for confidence intervals with matched comparisons \citep{MassonLoftus2003}.}
   \label{fig:majority_results}
\end{figure}

We chose the majority task because, in contrast to the parity task, we were uncertain if a restricted state representation would facilitate task performance. For the majority task of a given length $l$, the network needs to distinguish roughly $2l$ states. Collapsing them together is potentially dangerous: if the net does not keep exact count of the input sequence imbalance between 0's and 1's, it may fail.

As in the parity task, we tested both on novel binary sequences and training sequences with additive uniform noise. Figures~\ref{fig:majority_results}a,b show that neither the \RNN\ nor \RNNA\ beats \RNNSR\ for any sequence length on either test set. \RNNSR\ seems superior to the baseline \RNN\ for short novel sequences and long noisy sequences. For short noisy sequences, both architectures reach a ceiling. The only disappointment in this simulation is the lack of a difference for novel long sequences.

\subsubsection{Reber Grammar}

The Reber grammar \citep{reber1967}, shown in Figure~\ref{fig:majority_results}c, has long been a test case for artificial grammar learning \citep[e.g.,][]{hochreiter1997}. The task involves discriminating between strings that can and cannot be generated by the finite-state grammar. We generated positive strings by sampling from the grammar with uniform transition probabilities. Negative strings were generated from positive strings by substituting a single symbol for another symbol such that the resulting string is out-of-grammar. Examples of positive and negative strings are \textsc{btssxxttvpse} and \textsc{bptvpxt\color[rgb]{.75,0,0}s\color{black}pse}, respectively. Our networks used a one-hot encoding of the seven input symbols, $m=20$ tanh hidden units, $n=40$ attractor units, and $\sigma=0.25$. The number of training examples was varied from 50 to 800, always with 2000 test examples. Both the training and test sets were balanced to have an equal number of positive to negative strings. One hundred replications of each simulation was run.

Figure~\ref{fig:majority_results}d presents mean test set accuracy on the Reber grammar as a function of the number of examples used for training. As with previous data sets, \RNNSR\ outperforms the baseline \RNN, which in turn outperforms \RNNA.

\subsubsection{Symmetry Task}

The symmetry task involves detecting symmetry in fixed-length symbol strings such as \textsc{acafbbfaca}. This task is effectively a memory task for an RNN because the first half of the sequence must be retained to compare against the second half. We generated strings of length $2s+f$, where $s$ is the number of symbols in the left and right sides and $f$ is the number of intermediate fillers. For $i\in\{1,...,s\}$, we generated symbols $S_i \in \{ \textsc{a}, \textsc{b}, ..., \textsc{h} \}$. We then formed a string $X$ whose elements are determined by $S$: $X_i = S_i$ for $i \in \{1, ..., s\}$, $X_i = \emptyset$ for $i \in \{ s+1, ..., s+f \}$, and $X_i = S_{2s+f+1-i}$ for $i \in \{ s+f+1, ..., 2s+f\}$. The filler $\emptyset$ was simply a unique symbol. Negative cases were generated from a randomly drawn positive case by either exchanging two adjacent distinct non-null symbols, e.g., \textsc{acafb}\textsc{b\color[rgb]{.75,0,0}af\color{black}ca}, or substituting a single symbol with another, e.g., \textsc{a\color[rgb]{.75,0,0}h\color{black}afbbfaca}.
Our training and test sets had an equal number of positive and negative examples, and the negative examples were divided equally between the sequences with exchanges and substitutions. 

\begin{figure}[bt]
    \begin{center}
    \includegraphics[width=3in]{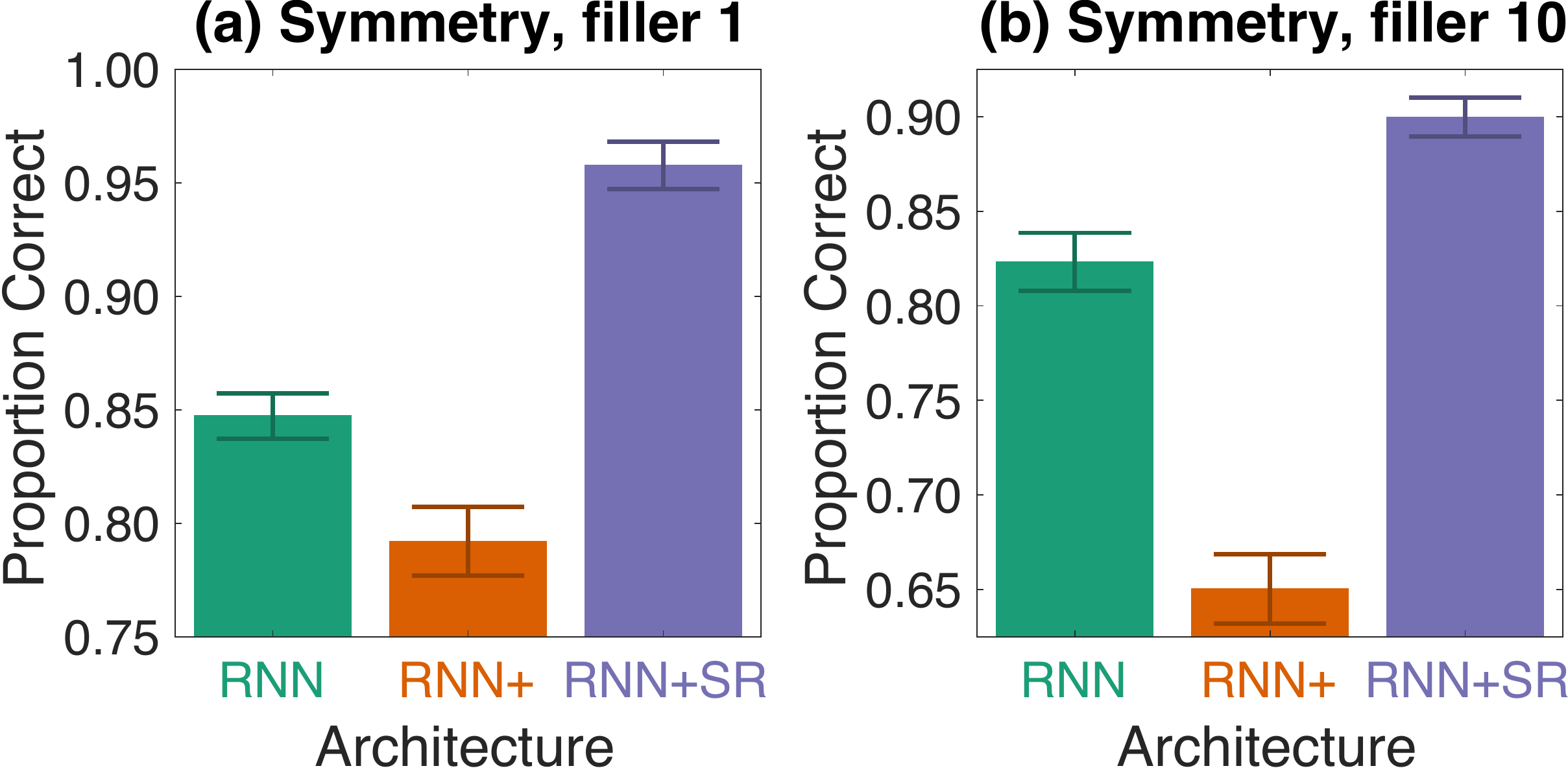} 
    \end{center}
    \caption{(a) Symmetry task with $f=1$ filler; (b) Symmetry task with $f=10$ filler. Error bars indicate $\pm1$ SEM.
    %, based on a correction for confidence intervals with matched comparisons \citep{MassonLoftus2003}
    }
   \label{fig:symmetry_results}
%\end{SCfigure*}
\end{figure}

We trained on 5000 examples and tested on an additional 2000, with the half sequence having length $s=5$ and with an $f=1$ or $f=10$ slot filler. The longer filler 
%requires that the memory of the input be maintained and it also 
makes temporal credit assignment more challenging. As shown in Figures~\ref{fig:symmetry_results}, \RNNSR\  obtains as much as a 70\% reduction in test error over either \RNN\ or \RNNA.

%sequences are 11 long, 5 symbols from {A,B,...H}
%rnn                        0.8474 0.8752 0.0138
%rnna                        0.7922 0.8520 0.0242
%sdprnn                     0.9576 0.9650 0.0048

%\begin{SCfigure*}[1.11111][b!]

\subsection{Language Modeling}

Turning to a second use of state reification, we explored whether the technique could be used to \emph{detect} when the hidden state has wandered from the training manifold. Our experiment was performed with an RNN language model that generates word predictions as output. The model can be run in generative mode by sampling from the output distribution and feeding it back to the input. This free-running mode often produces wide divergences from training \citep{Bengioetal2015}, because during training, teacher forcing ensures that the model's input sequence is a valid (observed) word sequence. The divergence increases as the sequence progresses.

Our experiment studies if state reification can detect when it has been given outputs from its own model (sampling mode) when trained using ground truth input sequences.  We trained a language model on the standard Text8 dataset, which is derived from Wikipedia articles.  We trained a single-layer LSTM with 1000 units at the character-level, and included DAE state-reification between the hidden states and the output on each time step.  
In a given sequence, following 50 sampling steps, the state reification layers had a reconstruction error on average 103\% of the teacher forcing reconstruction error.  Following 180 sampling steps, this value increased to 112\%.  Following 300 sampling steps this value increased even further to 134\%.  These results provide clear evidence that the outputs move off of the manifold with more sampling steps, and that this is effectively measured by state reification.

\subsection{Adversarial Training}
% MAYBE THIS SHOULD GO IN THE ADVERSARIAL SECTION
%We argue that applying the DAEs on the hidden layers---as opposed to the raw input signal---facilitates learning, while providing a stronger protection from adversarial attacks. 

In this section, we turn to a third application of state reification: obtaining networks robust to adversarial attacks, which consist of making small changes to input patterns that alter the predicted class. For image processing, the modulations of the input images can be small enough that they are unnoticeable to the human
eye; the modulations can be so robust that even when captured through a camera, they change the predicted class with high probability \citep{brown2017patch}. Such \emph{adversarial examples}~\citep{goodfellow2014adv} can be found via gradient-based methods \citep{szegedy2013adv,goodfellow2014adv}. 

% COMMENT FROM FORTIFIED PAPER: someone who knows more about adv examples literature should fix this paragraph
% Alex: I think people wouldn't agree that these are good defenses.  
Defenses proposed against adversarial examples include feature squeezing~\citep{xu2017squeeze}, adapted encoding of the input~\citep{buckman2018thermometer}, and distillation-related approaches~\citep{papernot2015distill}.  Many have been shown to be providing the illusion of defense by lowering the quality of the gradient signal, without actually providing improved robustness \citep{athalye2018obfuscate}.  One defense that is resilient to this \emph{obfuscated-gradient problem} is adversarial training \citep{madry2017adv}.  Adversarial training consists of augmenting the dataset with adversarial examples and training the model's predictions to be unchanged by the adversarial noise.

However, a major challenge of incorporating adversarial training is that adversarial robustness is often dramatically worse on test data as compared to train data, suggesting difficulty in generalization \citep{schmidt2018moredata}.  For this reason we explored the possibility of improving the performance of adversarial training by using state reification.  

Adversarial training is a flexible procedure and can be used with any adversarial attack.  For our investigation, we looked at the multi-step \emph{projected gradient descent} (\emph{PGD}) attack \citep{madry2017adv}.  We used an $l^{\infty}$ attack with $\epsilon$ ranging from $0.03$ to $0.3$ and number of iterations ranging from $7$ to $200$.  
%$\ell_\infty$-bounded adversaries via the following gradient based perturbation.
%\begin{align}
%\widetilde{x} = x + \varepsilon \operatorname{sgn}(\nabla_x %\loss(\theta,x,y)).
%\end{align}
The PGD attack ~\citep{madry2017adv}, also referred to as FGSM$^k$, is a multi-step extension of the Fast Gradient Sign Method (FGSM) ~\cite{goodfellow2014adv} attack.  The PGD attack is characterized as follows:
\begin{align}
x^{t+1} &= \Pi_{x+\hood} \left( x^t +
\alpha\operatorname{sgn}(\nabla_x \mathcal{L}_\mathrm{task}(x,y))\right) 
\end{align}
initialized with $x^0$ as the clean input $x$ and with the corrupted input $\widetilde{x}$ as the last step in the sequence.  $\Pi$ refers to the projection operator, which in this context means projecting the adversarial example back onto the region within an $\epsilon$ radius of the original data point after each step in the adversarial attack.  

%Our work differs from the approaches using generative models in the input space in that we instead employ state reification of the learned hidden representations, which projects off-manifold states back to the manifold and ought to make adversarial robustness transfer better to inputs from the test data.  

%which had two terms, to incorporate two dual terms. To explain, we start by identifying original training examples, $(x,y)$, for which an adversarial example can be identified. The adversarial example is a pair $(\widetilde{x},\widetilde{y}$ such that $x^{(i)}$ and $\widetilde{x}$ are similar (more on the notion of similarity shortly) and the resulting class $\widetilde{y}$ differs from the target class, $y$. The adversarial input produces a hidden state,
%$\widetilde{h}$, which we would like to project to the hidden state of the original input, $h$, via
%an adversarial reconstruction term:
%\begin{equation}
%\mathcal{L}_\mathrm{rec}(\widetilde{h},h) = 
%   \left(
%      \left\Vert r_\theta \left(\widetilde{h} + a \right) - h \right\Vert _2^2
%   \right),
%   \label{eqn:reif_loss}
%\end{equation}
%where $a\sim\mathbb{N}(0,\sigma^2)$ and $r_\theta$ is the reification function as before.
To apply state reification to adversarial training, we modified our original state-reification training loss (Equation~\ref{eqn:comb_loss}) with the standard adversarial training loss to encourage the network to not misclassify the adversarial example, yielding a combined loss for a given
example $(x,y)$ with an adversarial counterpart $\widetilde{x}$:
\begin{equation*}
    \mathcal{L} =
    \mathcal{L}_\mathrm{task}(x,y) + \mathcal{L}_\mathrm{task}(\widetilde{x},y) +
    \lambda_\mathrm{rec} \sum_{i \in S} \mathcal{L}_\mathrm{rec}^i (h_i)
\end{equation*}
where $S$ is the set of one or more hidden layers to which reification is applied, and the coefficient $ \lambda_\mathrm{rec}\ge0$ can be tuned to control the degree of reification.  Because we potentially apply reification to multiple layers, we replaced the AN of our earlier simulations with the simpler DAE.

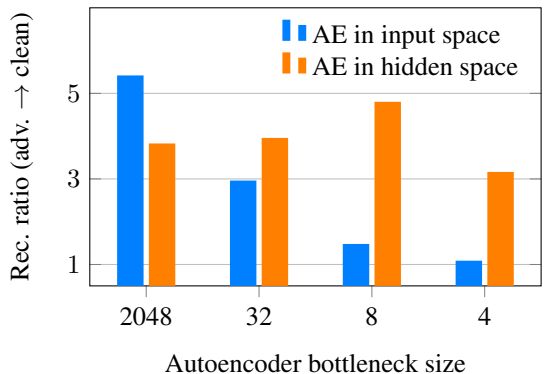
\begin{figure}[tb]
\centering
\begin{tikzpicture}
	\begin{axis}[
	    xmin=-0.5, xmax=3.5,
	    ymin=0.5, ymax=7,
	    scale only axis,
	    width=6cm, height=3.7cm,
		xlabel=Autoencoder bottleneck size,
		ylabel=Rec. ratio (adv. $\rightarrow$ clean),
		xtick={0,1,2,3},
		ytick={1,3,5},
    	xticklabels={2048, 32, 8, 4},
    	xtick pos=left,
    	ybar, bar width=1em,
    	ymajorgrids,
    	legend cell align={left},
    	legend style={draw=none},
    	x label style={at={(axis description cs:0.5,-0.07)},anchor=north},
        y label style={at={(axis description cs:0.1,.5)},anchor=south},
		]
	\addplot[draw=none,fill=blue!50!cyan,draw opacity=0] coordinates {
	(0, 0.0336/.0062) (1, 0.0468/0.0158) (2, 0.0570/0.0385) (3, 0.0589/0.0541)
	};
	\addplot[draw=none,fill=orange,draw opacity=0] coordinates {
	(0, 0.0157/.0041) (1, 0.0669/.0169) (2, 0.4548/0.0947) (3, 7.4066/2.3413)
	};
	\legend{
	    AE in input space,
	    AE in hidden space};
	\end{axis}
\end{tikzpicture}

\caption{Direct experimental evidence that reification is easier in hidden layers than the input: we added denoising autoencoders with different capacities to MLPs trained on MNIST, and display the value of the total reconstruction errors for adversarial examples divided by the total reconstruction errors for clean examples. A high value indicates success at detecting adversarial examples.  Our results support the central motivation for state reification: that off-manifold points can much more easily be detected in the hidden space (as seen by the relatively constant ratio for the autoencoder in hidden space) and are much harder to detect in the input space (as seen by this ratio rapidly falling to zero as the input-space autoencoder's capacity is reduced}
\label{fig:mnistae}
\end{figure}

%An important question is: how do fortified networks compare to DAEs trained in the `pixel-space`? Autoencoders are capable of handling arbitrary data distributions in principle (with enough capacity, training, etc). 
% (I'm commenting the above part out because it appears to be a repetition of what is said below.
%The denoising process can be done in `pixel-space' (that is, applying a DAE on the input data $x$) or in the hidden space. Why do we work in the learned latent space instead?
We have discussed advantages to performing reification in the hidden space instead of the input space, but the question of where exactly reification should be performed in a deep net remains unanswered: just the final hidden layer? Every hidden layer? We outline two important considerations regarding this issue.  On the one hand, identifying states that are off-manifold or close to the margin is easier in the deeper hidden layers (see Figure~\ref{fig:mnistae}, which we explain shortly).  On the other hand, the states in the deeper hidden layers may already look non-adversarial due to the effect of the adversarial perturbations in the shallower layers.  While we are not aware of any formal study of this phenomenon, it is clearly possible. (Imagine, for example, state reification performed on the output from the classifier softmax, which could only identify unnatural combinations of class probabilities.)  Given these opposing concerns, we argue for the inclusion of reification at multiple stages of the network, very much analogous to the inclusion of reification at each time step of the recurrent net in our previous simulations.

We collected  experimental evidence that more directly supports our decision to perform state reification at many levels of representation.  We constructed FGSM adversarial examples ($\epsilon=0.3$) on small MNIST fully-connected networks trained normally.  As Figure~\ref{fig:mnistae} shows, we found that detecting adversarial examples by reconstruction error is possible both in input and hidden layers, but could be performed by much smaller autoencoders via the hidden layers.

Tables~\ref{tb:cifar_cnn} and \ref{tb:cifar_resnet} present results applying state reification on CIFAR10 using non-ResNet and ResNet convolutional nets (CNNs), respectively. Substantially better test-set adversarial robustness is attained via adversarial training when done in conjunction with state reification, evaluated on a wide range of $\epsilon$ values ($0.03$ to $0.3$) and number of attack steps (7 to 200).  

\citet{athalye2018obfuscate} suggest that models which introduce components with noisy or unreliable gradients can reduce the quality of gradient-based attacks.  To test this hypothesis, they introduced \emph{backward-pass differentiable approximation}, where the attack treats the ``reconstructor'' (in our case, the DAE) as the identity function when computing gradients for the attack.  Our results showed that bypassing the DAE substantially reduced the strength of the attack, resulting in an increase in PGD accuracy to 67.1\% from 40.1\% (higher accuracy implies a weaker attack).  Additionally, we ran a noiseless attack in which the forward and backward passes were performed without noise. This change strengthened the attack: PGD accuracy rises to 40.1\% from 38.2\% (lower accuracy implies a stronger attack), but we note that this is much less than the overall gap between state reification and the same-capacity baseline, suggesting that adding noise did partially obfuscate gradients, but not to such a degree as to nullify the improvements from state reification.  

{\renewcommand{\arraystretch}{1.0}
\begin{table}[ht!]
\centering
\caption{CIFAR-10 PGD Results with (non-ResNet) CNNs.  In these experiment we apply state reification (with single hidden layer convolutional autoencoders) following each convolutional layer.  Both experiments were run for 200 epochs and with all hyperparameters and architecture kept the same with the exception of state reification being added.  We considered different types of baselines: \CNN\ means we simply remove state reification. \CNNA\ means that we added extra layers with the same number of units to match the capacity added by state reification.  The three blocks of results use slightly different architectures for the CNNs and are thus not directly comparable.  All models reported were trained with adversarial training with a PGD attack.  Note that higher PGD accuracy indicates a stronger defense.}
\label{tb:cifar_cnn}
\begin{tabular}{cccrrr} %\hline
\toprule
%Attack Type & \shortstack{PGD\\Steps} &
%\shortstack{Attack\\Epsilon} & \multicolumn{3}{c}{PGD Accuracy} \\
%&&&\CNN  & \CNNA & \CNNSR \\
Attack& PGD & Attack & \multicolumn{3}{c}{PGD Accuracy} \\
Type & Steps & Epsilon &\CNN  & \CNNA & \CNNSR \\
\midrule
%Baseline - extra activations & Normal & 7 & 0.03 & 38.1 \\
% MM to ALEX: What is this next line?
%\CNNSR & Normal & 7 & 0.03 & 43.3 \\
Normal & 7 & 0.03 & 33.0 & 34.2 & 45.0 \\
Normal & 50 & 0.03 & 31.6 & 32.5 & 42.1\\
Normal & 200 & 0.03 & 31.4 & 32.2 & 41.5\\
%\CNN & Normal & 7 & 0.03 & 33.0 \\
%\CNN & Normal & 50 & 0.03 & 31.6 \\
%\CNN & Normal & 200 & 0.03 & 31.4 \\
%\CNNA & Normal & 7 & 0.03 & 34.2 \\
%\CNNA & Normal & 50 & 0.03 & 32.5 \\
%\CNNA & Normal & 200 & 0.03 & 32.2 \\
%\CNNSR & Normal & 7 & 0.03 & 45.0 \\
%\CNNSR & Normal & 50 & 0.03 & 42.1 \\
%\CNNSR & Normal & 200 & 0.03 & 41.5 \\
\midrule
Normal & 100 & 0.03 && 35.3 & 39.2\\
Normal & 100 & 0.04 && 24.8 & 28.0\\
Normal & 100 & 0.06 && 14.3 &15.6\\
Normal & 100 & 0.08 && 12.0 & 13.0\\
Normal & 100 & 0.10 && 11.7 & 12.9\\
Normal & 100 & 0.20 && 10.2 & 11.3\\
Normal & 100 & 0.30 && 8.4 & 9.6 \\
%\CNNA & Normal & 100 & 0.03 & 35.3 \\
%\CNNA & Normal & 100 & 0.04 & 24.8 \\
%\CNNA & Normal & 100 & 0.06 & 14.3 \\
%\CNNA & Normal & 100 & 0.08 & 12.0 \\
%\CNNA & Normal & 100 & 0.1 & 11.7 \\
%\CNNA & Normal & 100 & 0.2 & 10.2 \\
%\CNNA & Normal & 100 & 0.3 & 8.40 \\
%\CNNSR & Normal & 100 & 0.03 & 39.2 \\
%\CNNSR & Normal & 100 & 0.04 & 28.0 \\
%\CNNSR & Normal & 100 & 0.06 & 15.6 \\
%\CNNSR & Normal & 100 & 0.08 & 13.0 \\
%\CNNSR & Normal & 100 & 0.1 & 12.9 \\
%\CNNSR & Normal & 100 & 0.2 & 11.3 \\
%\CNNSR & Normal & 100 & 0.3 & 9.60 \\
\midrule
%\CNNA & Normal & 100 & 0.03 & 33.4 \\
%\CNNSR & Normal & 100 & 0.03 & 40.1 \\
Normal & 100 & 0.03 & & 33.4 & 40.1 \\
%\CNNSR & \makecell[l]{Noiseless\\Attack} & 100 & 0.03 & 38.2 \\
\makecell[l]{Noiseless\\Attack} & 100 & 0.03 & & & 38.2 \\
%\CNNSR & \makecell[l]{BPDA,\\Skip-DAE} & 100 & 0.03 & 67.1 \\
\makecell[l]{BPDA,\\Skip-DAE} & 100 & 0.03 & & & 67.1 \\
\bottomrule
\end{tabular}
\vspace{-1em}
\end{table}
}

{\renewcommand{\arraystretch}{1.0}
\begin{table}[ht]
\centering
\caption{CIFAR-10 PGD Results with two powerful ResNet architectures: PreActResNet18 \citep{he2016preact} and WideResNet28-10 \citep{zagoruyko2016wrn}.  In this experiment we used a single state reification layer following the 2nd resblock (\emph{ResNet-SR}); the baseline consists of the same network with the state reification removed (\emph{ResNet}).  Both experiments were run for 200 epochs and with all hyperparameters and architecture kept the same.  Note that higher PGD accuracy indicates a stronger defense.\\  }
\label{tb:cifar_resnet}
\begin{tabular}{ccc} %\hline
\toprule
 & \multicolumn{2}{c}{PGD Accuracy (20 steps)} \\
Model&baseline&\sc{sr}\\
\midrule
PreActResNet18 & 37.87 & 39.20 \\
WideResNet28-10  & 43.28 & 44.06 \\
\bottomrule
\end{tabular}
\vspace{-1em}
\end{table}
}

\section{Related Work}

State reification seems related to several recent papers with a cognitive science focus.
\citet{AndreasKleinLevine2017} proposed a model that efficiently learns new concepts and control policies by operating in a linguistically constrained representational space. The space is obtained by pretraining on a language task, and this pretraining imposes structure on subsequent learning. One can view reification as imposing similar structure, although the bias comes not from a separate task or data set, but from representations already learned for the primary task. Related to language, the \emph{consciousness prior} of \citet{bengio2017} suggests a potential role of operating in a reduced or simplified representational space. Bengio conjectures that the high dimensional state space of the brain is unwieldy, and a restricted representation that selects some information at the expense of other may facilitate rapid learning and efficient inference. For related ideas, also see \citet{hinton1990}.

On the subject of our experimental results on adversarial robustness, the observation that adversarial examples often consist of points off  the data manifold and that deep networks may not generalize well to these points motivated several authors  to consider the use of the generative models as a defense against adversarial attacks \citep{gu2014robust,ilyas2017robust,samangouei2018defensegan,liao2017defense}.  \citet{ilyas2017robust,gilmer2018sphere} also showed the existence of adversarial examples which lie on the data manifold, and \citet{ilyas2017robust} showed that training against adversarial examples forced to lie on the manifold is an effective defense.  Our method shares a closely related motivation to these prior works, with a key difference being that we propose to consider the manifold in the space of learned representations, not the manifold directly in the visible space.  One motivation for this is that the learned representations have a simpler statistical structure \citep{bengio2012mix}, which makes the task of modeling this manifold and detecting unnatural points much simpler.  Learning the distribution directly in the visible space is still very difficult (even state of the art models fall short of real data on metrics like Inception Score) and requires a high capacity model.  Additionally, working in the space of learned representations allows for the use of a relatively simple generative model, in our case a small denoising autoencoder.  Finally, another important difference is that we always use state reification together with adversarial training.  

% Removed for space
%\citet{kannan2018logit} proposed a method which involves matching the logit (pre-softmax outputs) values for the original samples with the logit values resulting from adversarial examples.  

Denoising Feature Matching \citep{warde2017improving} proposed to train a denoising autoencoder in the hidden states of the discriminator in a generative adversarial network.  The generator's parameters are then trained to make the reconstruction error of this autoencoder small.  This has the effect of encouraging the generator to produce points which are easy for the model to reconstruct, which will include true data points.  Both this and state reification use a learned denoising autoencoder in the hidden states of a network.  A major difference is that the denoising feature matching work focused on generative adversarial networks and tried to minimize reconstruction error through a learned generator network, whereas our approach targets the adversarial examples problem.  Additionally, our objective encourages the output of the DAE to denoise adversarial examples so as to point back to the hidden state of the original example, which is different from the objective in the denoising feature matching work, which encouraged reconstruction error to be low on states from samples from the generator network.

MagNet \citep{MengChen2017} also proposed a method using autoencoders in the input space of a deep network to detect adversarial examples and ``reform'' them back to the input space.  Their work differs from our approach in two critical ways.  First, our method uses denoising autoencoders at several levels of representation, whereas MagNet \citep{MengChen2017} only operated in the input space.  Second, our method is used together with adversarial training and is motivated primarily from the perspective of improving generalization in adversarial training.  Many methods that have used autoencoders by themselves as a defense against adversarial examples are successful only when the autoencoder is ignored during the attack \citep{athalye2018obfuscate}; however, with state reification, we are able to improve robustness even when the autoencoder is used for the attack.  In Table~\ref{tb:cifar_cnn}, we also present various alternative attacks that skip the autoencoder or don't inject noise, and found that robustness was preserved in all cases. 

\citet{gilmer2018sphere} studied the existence of adversarial examples in the task of classifying between two hollow concentric shells.  Intriguingly, they prove and construct adversarial examples which lie on the data manifold (although \citeauthor{ilyas2017robust}, \citeyear{ilyas2017robust}, also looked for such examples experimentally using GANs).  The existence of such on-manifold adversarial examples demonstrates that a simplified version of our model trained with only $\mathcal{L}_\mathrm{rec}$ and not adversarial training could not protect against all adversarial examples.  However, combined with adversarial training, state reification may still help with on-manifold adversarial examples as well by mapping the hidden state back to regions where the model performs well.

\section{Discussion}

% MM 1/23: Alex prefers removing the first 2 sentences. I like. Other opinions? We don't want to suggest in any way that what we're doing is attaining noise robustness!
Noise robustness is a highly desirable property in neural networks. When a neural net performs well, it naturally exhibits a sort of noise suppression: activation in a layer is relatively invariant to noise injected at lower layers \citep{Aroraetal2018}. We described a method, state reification, which has the explicit objective of attaining robustness to unfamiliar variation, and we demonstrated that state reification helps neural nets to generalize better, especially when labeled data are sparse, and also helps overcome the challenge of achieving robust generalization with adversarial training. We also described two different implementation substrates for state reification, one using attractor nets and the other denoising autoencoders. We suspect that other kinds of unsupervised learning mechanisms that perform representation compression and density estimation will work as well if not better, especially those with explicit probabilistic underpinnings.

Our aim has been to show that state reification is an idea with breadth---over the quite disparate domains of symbolic sequence recognition and generation tasks and adversarial robustness. Although state reification appears to have some practical uses, more basic research is needed to understand how neural nets perform in regions of hidden state space outside the training manifold. More broadly, state reification addresses an issue that is often neglected in deep learning: how to build robust models given that internal state spaces are continuous, high dimensional, and often unbounded.  The human brain has solved this problem, and artificial intelligence needs to do so as well.

\clearpage
\bibliographystyle{apalike}
\bibliography{icml2019}

\begin{thebibliography}{}

\bibitem[Alain et~al., 2012]{alain2012dae}
Alain, G., Bengio, Y., and Rifai, S. (2012).
\newblock Regularized auto-encoders estimate local statistics.
\newblock {\em CoRR}, abs/1211.4246.

\bibitem[Andreas et~al., 2017]{AndreasKleinLevine2017}
Andreas, J., Klein, D., and Levine, S. (2017).
\newblock Learning with latent language.
\newblock {\em CoRR}, abs/1711.00482.

\bibitem[Arora et~al., 2018]{Aroraetal2018}
Arora, S., Ge, R., Neyshabur, B., and Zhang, Y. (2018).
\newblock Stronger generalization bounds for deep nets via a compression
  approach.
\newblock {\em CoRR}, abs/1802.05296.

\bibitem[{Athalye} et~al., 2018]{athalye2018obfuscate}
{Athalye}, A., {Carlini}, N., and {Wagner}, D. (2018).
\newblock {Obfuscated Gradients Give a False Sense of Security: Circumventing
  Defenses to Adversarial Examples}.
\newblock {\em ArXiv e-prints}.

\bibitem[Bengio et~al., 2015]{Bengioetal2015}
Bengio, S., Vinyals, O., Jaitly, N., and Shazeer, N. (2015).
\newblock Scheduled sampling for sequence prediction with recurrent neural
  networks.
\newblock In Cortes, C., Lawrence, N.~D., Lee, D.~D., Sugiyama, M., and
  Garnett, R., editors, {\em Advances in Neural Information Processing Systems
  28}, pages 1171--1179. Curran Associates, Inc.

\bibitem[Bengio, 2017]{bengio2017}
Bengio, Y. (2017).
\newblock The consciousness prior.
\newblock {\em CoRR}, abs/1709.08568.

\bibitem[Bengio et~al., 2013]{Bengio-Courville-Vincent-TPAMI2013}
Bengio, Y., Courville, A., and Vincent, P. (2013).
\newblock Representation learning: A review and new perspectives.
\newblock {\em {IEEE} Trans. Pattern Analysis and Machine Intelligence (PAMI)},
  35(8):1798--1828.

\bibitem[Bengio et~al., 2012]{bengio2012mix}
Bengio, Y., Mesnil, G., Dauphin, Y., and Rifai, S. (2012).
\newblock Better mixing via deep representations.
\newblock {\em CoRR}, abs/1207.4404.

\bibitem[Bengio et~al., 1994]{Bengio1994}
Bengio, Y., Simard, P., and Frasconi, P. (1994).
\newblock Learning long-term dependencies with gradient descent is difficult.
\newblock {\em Trans. Neur. Netw.}, 5(2):157--166.

\bibitem[Boll, 1979]{Boll1979}
Boll, S. (1979).
\newblock Suppression of acoustic noise in speech using spectral subtraction.
\newblock {\em IEEE Transactions on Acoustics, Speech, and Signal Processing},
  27(2):113--120.

\bibitem[{Brown} et~al., 2017]{brown2017patch}
{Brown}, T.~B., {Man{\'e}}, D., {Roy}, A., {Abadi}, M., and {Gilmer}, J.
  (2017).
\newblock {Adversarial Patch}.
\newblock {\em ArXiv e-prints}.

\bibitem[Carrara et~al., 2018]{Carraraetal2018}
Carrara, F., Becarelli, R., Caldelli, R., Falchi, F., and Amato, G. (2018).
\newblock Adversarial examples detection in features distance spaces.
\newblock http://www.nmis.isti.cnr.it/falchi/Draft/2018-ECCV-WOCM-Draft.pdf.

\bibitem[Craven and Shavlik, 1993]{CravenShavlik1993}
Craven, M.~W. and Shavlik, J.~W. (1993).
\newblock Learning symbolic rules using artificial neural networks.
\newblock In {\em Proceedings of the Tenth International Conference on Machine
  Learning}, pages 73--80. Morgan Kaufmann.

\bibitem[{Gilmer} et~al., 2018]{gilmer2018sphere}
{Gilmer}, J., {Metz}, L., {Faghri}, F., {Schoenholz}, S.~S., {Raghu}, M.,
  {Wattenberg}, M., and {Goodfellow}, I. (2018).
\newblock {Adversarial Spheres}.
\newblock {\em ArXiv e-prints}.

\bibitem[{Goodfellow} et~al., 2014]{goodfellow2014adv}
{Goodfellow}, I.~J., {Shlens}, J., and {Szegedy}, C. (2014).
\newblock {Explaining and Harnessing Adversarial Examples}.
\newblock {\em ArXiv e-prints}.

\bibitem[Gu and Rigazio, 2014]{gu2014robust}
Gu, S. and Rigazio, L. (2014).
\newblock Towards deep neural network architectures robust to adversarial
  examples.
\newblock {\em CoRR}, abs/1412.5068.

\bibitem[He et~al., 2016]{he2016preact}
He, K., Zhang, X., Ren, S., and Sun, J. (2016).
\newblock Identity mappings in deep residual networks.
\newblock {\em CoRR}, abs/1603.05027.

\bibitem[Hinton, 1990]{hinton1990}
Hinton, G.~E. (1990).
\newblock Mapping part-whole hierarchies into connectionist networks.
\newblock {\em Artificial Intelligence}, 46:47--75.

\bibitem[Hochreiter, 1998]{hochreiter1998}
Hochreiter, S. (1998).
\newblock The vanishing gradient problem during learning recurrent neural nets
  and problem solutions.
\newblock {\em Int. J. Uncertain. Fuzziness Knowl.-Based Syst.}, 6(2):107--116.

\bibitem[Hochreiter and Schmidhuber, 1997]{hochreiter1997}
Hochreiter, S. and Schmidhuber, J. (1997).
\newblock Long short-term memory.
\newblock {\em Neural Computation}, 9(8):1735--1780.

\bibitem[Hopfield, 1982]{Hopfield1982}
Hopfield, J.~J. (1982).
\newblock Neural networks and physical systems with emergent collective
  computational abilities.
\newblock {\em Proceedings of the national academy of sciences},
  79(8):2554--2558.

\bibitem[Hopfield, 1984]{Hopfield1984}
Hopfield, J.~J. (1984).
\newblock Neurons with graded response have collective computational properties
  like those of two-state neurons.
\newblock {\em Proceedings of the national academy of sciences},
  81(10):3088--3092.

\bibitem[{Ilyas} et~al., 2017]{ilyas2017robust}
{Ilyas}, A., {Jalal}, A., {Asteri}, E., {Daskalakis}, C., and {Dimakis}, A.~G.
  (2017).
\newblock {The Robust Manifold Defense: Adversarial Training using Generative
  Models}.
\newblock {\em ArXiv e-prints}.

\bibitem[Jacob~Buckman, 2018]{buckman2018thermometer}
Jacob~Buckman, Aurko~Roy, C. R. I.~G. (2018).
\newblock Thermometer encoding: One hot way to resist adversarial examples.
\newblock {\em International Conference on Learning Representations}.

\bibitem[Koiran, 1994]{Koiran1994}
Koiran, P. (1994).
\newblock Dynamics of discrete time, continuous state hopfield networks.
\newblock {\em Neural Computation}, 6(3):459--468.

\bibitem[Lee et~al., 2017]{LeeLeeLeeShin2017}
Lee, K., Lee, K., Lee, H., and Shin, J. (2017).
\newblock Training confidence-calibrated classifiers for detecting
  out-of-distribution samples.
\newblock {\em CoRR}, abs/1711.09325.

\bibitem[Lee et~al., 2018]{LeeLeeLeeShin2018}
Lee, K., Lee, K., Lee, H., and Shin, J. (2018).
\newblock A simple unified framework for detecting out-of-distribution samples
  and adversarial attacks.
\newblock In Bengio, S., Wallach, H., Larochelle, H., Grauman, K.,
  Cesa-Bianchi, N., and Garnett, R., editors, {\em Advances in Neural
  Information Processing Systems 31}, pages 7167--7177. Curran Associates, Inc.

\bibitem[{Liao} et~al., 2017]{liao2017defense}
{Liao}, F., {Liang}, M., {Dong}, Y., {Pang}, T., {Zhu}, J., and {Hu}, X.
  (2017).
\newblock {Defense against Adversarial Attacks Using High-Level Representation
  Guided Denoiser}.
\newblock {\em ArXiv e-prints}.

\bibitem[Liao et~al., 2016]{Liao2016}
Liao, R., Schwing, A., Zemel, R., and Urtasun, R. (2016).
\newblock Learning deep parsimonious representations.
\newblock In Lee, D.~D., Sugiyama, M., Luxburg, U.~V., Guyon, I., and Garnett,
  R., editors, {\em Advances in Neural Information Processing Systems 29},
  pages 5076--5084. Curran Associates, Inc.

\bibitem[{Madry} et~al., 2017]{madry2017adv}
{Madry}, A., {Makelov}, A., {Schmidt}, L., {Tsipras}, D., and {Vladu}, A.
  (2017).
\newblock {Towards Deep Learning Models Resistant to Adversarial Attacks}.
\newblock {\em ArXiv e-prints}.

\bibitem[Masson and Loftus, 2003]{MassonLoftus2003}
Masson, M. and Loftus, G. (2003).
\newblock Using confidence intervals for graphically based data interpretation.
\newblock {\em Canadian Journal of Experimental Psychology}, 57:203--220.

\bibitem[Meng and Chen, 2017]{MengChen2017}
Meng, D. and Chen, H. (2017).
\newblock Magnet: A two-pronged defense against adversarial examples.
\newblock In {\em Proceedings of the 2017 ACM SIGSAC Conference on Computer and
  Communications Security}, CCS '17, pages 135--147, New York, NY, USA. ACM.

\bibitem[Mozer, 2009]{Mozer2009}
Mozer, M.~C. (2009).
\newblock Attractor networks.
\newblock In Bayne, T., Cleeremans, A., and Wilken, P., editors, {\em Oxford
  Companion to Consciousness}, pages 88--89. Oxford University Press, Oxford
  UK.

\bibitem[Pang et~al., 2018]{PangDuDongZhu2018}
Pang, T., Du, C., Dong, Y., and Zhu, J. (2018).
\newblock Towards robust detection of adversarial examples.
\newblock In Bengio, S., Wallach, H., Larochelle, H., Grauman, K.,
  Cesa-Bianchi, N., and Garnett, R., editors, {\em Advances in Neural
  Information Processing Systems 31}, pages 4584--4594. Curran Associates, Inc.

\bibitem[Papernot et~al., 2015]{papernot2015distill}
Papernot, N., McDaniel, P.~D., Wu, X., Jha, S., and Swami, A. (2015).
\newblock Distillation as a defense to adversarial perturbations against deep
  neural networks.
\newblock {\em CoRR}, abs/1511.04508.

\bibitem[Reber, 1967]{reber1967}
Reber, A.~S. (1967).
\newblock Implicit learning of artificial grammars.
\newblock {\em Verbal learning and verbal behavior}, 5:855--863.

\bibitem[Samangouei et~al., 2018]{samangouei2018defensegan}
Samangouei, P., Kabkab, M., and Chellappa, R. (2018).
\newblock Defense-gan: Protecting classifiers against adversarial attacks using
  generative models.
\newblock In {\em International Conference on Learning Representations},
  volume~9.

\bibitem[Schmidt et~al., 2018]{schmidt2018moredata}
Schmidt, L., Santurkar, S., Tsipras, D., Talwar, K., and Madry, A. (2018).
\newblock Adversarially robust generalization requires more data.
\newblock {\em CoRR}, abs/1804.11285.

\bibitem[Siegelmann, 2008]{Siegelmann2008}
Siegelmann, H.~T. (2008).
\newblock Analog-symbolic memory that tracks via reconsolidation.
\newblock {\em Physica D: Nonlinear Phenomena}, 237(9):1207 -- 1214.

\bibitem[Simard et~al., 1992]{Simard1992}
Simard, P., Victorri, B., LeCun, Y., and Denker, J. (1992).
\newblock Tangent prop - a formalism for specifying selected invariances in an
  adaptive network.
\newblock In Moody, J.~E., Hanson, S.~J., and Lippmann, R.~P., editors, {\em
  Advances in Neural Information Processing Systems 4}, pages 895--903.
  Morgan-Kaufmann.

\bibitem[{Szegedy} et~al., 2013]{szegedy2013adv}
{Szegedy}, C., {Zaremba}, W., {Sutskever}, I., {Bruna}, J., {Erhan}, D.,
  {Goodfellow}, I., and {Fergus}, R. (2013).
\newblock {Intriguing properties of neural networks}.
\newblock {\em ArXiv e-prints}.

\bibitem[Warde-Farley and Bengio, 2017]{warde2017improving}
Warde-Farley, D. and Bengio, Y. (2017).
\newblock Improving generative adversarial networks with denoising feature
  matching.
\newblock In {\em International Conference on Learning Representations 2017
  (Conference Track)}.

\bibitem[Xu et~al., 2017]{xu2017squeeze}
Xu, W., Evans, D., and Qi, Y. (2017).
\newblock Feature squeezing: Detecting adversarial examples in deep neural
  networks.
\newblock {\em CoRR}, abs/1704.01155.

\bibitem[Zagoruyko and Komodakis, 2016]{zagoruyko2016wrn}
Zagoruyko, S. and Komodakis, N. (2016).
\newblock Wide residual networks.
\newblock {\em CoRR}, abs/1605.07146.

\bibitem[Zheng et~al., 2016]{Zheng2016}
Zheng, S., Song, Y., Leung, T., and Goodfellow, I. (2016).
\newblock Improving the robustness of deep neural networks via stability
  training.
\newblock In {\em Proceedings of the IEEE Conference on Computer Vision and
  Pattern Recognition}, pages 4480--4488.

\end{thebibliography}

%\clearpage
%\appendix
%\input{appendix.tex}

\end{document}